\documentclass{article}

\usepackage[accepted]{include/aistats/aistats2024}

\usepackage[utf8]{inputenc} %
\usepackage[T1]{fontenc}    %
\usepackage{hyperref}       %
\usepackage{url}            %
\usepackage{booktabs}       %
\usepackage{amsfonts}       %
\usepackage{nicefrac}       %
\usepackage{microtype}      %
\usepackage{xcolor}         %

\usepackage{graphicx}
\usepackage{subfig}

\usepackage{include/titletoc}
\usepackage{amsmath}
\usepackage{amssymb}
\usepackage{natbib}

\usepackage{multirow}
\usepackage{array}
\usepackage{multicol}
\usepackage{tabularx}
\usepackage{tabularray}
\UseTblrLibrary{booktabs}
\usepackage[inline]{enumitem}

\usepackage[utf8]{inputenc}
\usepackage{microtype}
\usepackage{graphicx}
\usepackage{booktabs} %
\usepackage{titletoc}
\usepackage{hyperref}

\usepackage{amsmath}
\usepackage{amssymb}
\usepackage{bbm} 
\usepackage{bm}
\usepackage{verbatim}
\usepackage{float}
\usepackage{color,soul}
\usepackage{enumitem}
\usepackage{mathtools}
\usepackage{hhline}
\usepackage[title]{appendix}
\usepackage[nameinlink]{cleveref} %
\usepackage{tabularx}
\usepackage{tikz}
\usepackage{wrapfig,booktabs}
\usepackage{xspace}
\usepackage{cancel}
\usepackage{sidecap}

\graphicspath{ {../figures/} }

\DeclarePairedDelimiterX{\infdivx}[2]{(}{)}{%
  #1\;\delimsize\|\;#2%
}

\crefname{appsec}{appendix}{appendices}
\Crefname{appsec}{Appendix}{Appendices}

\definecolor{mydarkblue}{rgb}{0,0.08,0.45}
\hypersetup{ %
    pdftitle={},
    pdfauthor={},
    pdfsubject={Proceedings of the International Conference on Machine Learning 2020},
    pdfkeywords={},
    pdfborder=0 0 0,
    pdfpagemode=UseNone,
    colorlinks=true,
    linkcolor=mydarkblue,
    citecolor=mydarkblue,
    filecolor=mydarkblue,
    urlcolor=mydarkblue,
    pdfview=FitH
}

\usetikzlibrary{shapes.geometric, arrows, bayesnet, calc, positioning}

\newcommand{\dee}{\,\textrm{d}}

\newcommand{\calF}{\mathcal{F}}

\newcommand{\calN}{\mathcal{N}}

\newcommand{\calL}{\mathcal{L}}

\newcommand{\closer}[3]{{\kern-#1ex{#2}\kern-#3ex}}

\DeclareMathOperator*{\argmax}{arg\,max}

\mathchardef\mhyphen="2D

\usepackage{titletoc}
\usepackage{algorithm}
\usepackage{algorithmic}
\usepackage{colortbl}

\definecolor{azure}{rgb}{0.0, 0.5, 1.0}
\definecolor{airforceblue}{rgb}{0.36, 0.54, 0.66}
\definecolor{darkgreen}{rgb}{0.0, 0.2, 0.13}

\newcommand\defines{\,\dot{=}\,}

\newcommand{\map}{\textsc{map}\xspace}

\newcommand{\vbar}{\,|\,}

\newcommand{\calX}{\mathcal{X}}
\newcommand{\calY}{\mathcal{Y}}
\newcommand{\calD}{\mathcal{D}}

\newcommand{\pms}[1]{\ensuremath{{\scriptstyle\pm #1}}}

\usepackage{standalone}
\usepackage{tikz}
\usepackage{pgfplots}
\usepackage{pgf}
\usetikzlibrary{calc}
\usetikzlibrary{positioning}
\usetikzlibrary{angles,quotes}
\usetikzlibrary{backgrounds}
\usetikzlibrary{fit}
\usetikzlibrary{arrows}
\usetikzlibrary{arrows.meta}
\usetikzlibrary{shapes.symbols}
\usetikzlibrary{shadings}
\usetikzlibrary{shapes}
\usetikzlibrary{fadings}
\usetikzlibrary{bayesnet}
\usetikzlibrary{matrix}
\usetikzlibrary{plotmarks}
\usetikzlibrary{intersections}
\usetikzlibrary{pgfplots.fillbetween}
\pgfplotsset{compat=1.14}
\usepackage{siunitx}

\usepackage{colortbl}
\definecolor{mediumgray}{gray}{0.7}
\definecolor{lightgray}{gray}{0.76}
\definecolor{lightlightgray}{gray}{0.9}
\definecolor{C1}{HTML}{1F77B4}
\definecolor{C2}{HTML}{FF7F0E}
\definecolor{C3}{HTML}{2CA02C}
\definecolor{C4}{HTML}{D62728}
\definecolor{C5}{HTML}{9467BD}
\colorlet{C1light}{C1!70!white}
\colorlet{C2light}{C2!70!white}
\colorlet{C3light}{C3!70!white}
\colorlet{C4light}{C4!70!white}
\colorlet{C5light}{C5!70!white}
\colorlet{C1lighter}{C1!50!white}
\colorlet{C2lighter}{C2!50!white}
\colorlet{C3lighter}{C3!50!white}
\colorlet{C4lighter}{C4!50!white}
\colorlet{C5lighter}{C5!50!white}
\colorlet{C1vlight}{C1!20!white}
\colorlet{C2vlight}{C2!20!white}
\colorlet{C3vlight}{C3!20!white}
\colorlet{C4vlight}{C4!20!white}
\colorlet{C5vlight}{C5!20!white}
\colorlet{linkcolor}{violet}

\usepackage{tcolorbox}

\usepackage{xcolor}

\definecolor{mydarkblue}{rgb}{0,0.08,0.45}
\definecolor{ourmethodblue}{rgb}{0,0.447,0.698}
\definecolor{ourmethodlblue}{rgb}{0.651, 0.808, 0.890}
\definecolor{evenlighterblue}{RGB}{213,241,255}

\hypersetup{ %
    pdftitle={},
    pdfauthor={},
    pdfsubject={},
    pdfkeywords={},
    pdfborder=0 0 0,
    pdfpagemode=UseNone,
    colorlinks=true,
    linkcolor=mydarkblue,
    citecolor=mydarkblue,
    filecolor=mydarkblue,
    urlcolor=mydarkblue,
    pdfview=FitH
}

\begin{document}

\twocolumn[

\aistatstitle{Mind the GAP: Improving Robustness to\\Subpopulation Shifts with Group-Aware Priors}

\aistatsauthor{Tim G. J. Rudner
\And Ya Shi Zhang \And Andrew Gordon Wilson \And Julia Kempe 
}

\aistatsaddress{New York University \And New York University \And New York University \And New York University\\Meta FAIR} ]

\vspace*{-15pt}
\begin{abstract}
Machine learning models often perform poorly under subpopulation shifts in the data distribution. Developing methods that allow machine learning models to better generalize to such shifts is crucial for safe deployment in real-world settings. In this paper, we develop a family of group-aware prior (GAP) distributions over neural network parameters that explicitly favor models that generalize well under subpopulation shifts. We design a simple group-aware prior that only requires access to a small set of data with group information and demonstrate that training with this prior yields state-of-the-art performance---even when only retraining the final layer of a previously trained non-robust model. Group aware-priors are conceptually simple, complementary to existing approaches, such as attribute pseudo labeling and data reweighting, and open up promising new avenues for harnessing Bayesian inference to enable robustness to subpopulation shifts.
\end{abstract}

\vspace*{-7pt}
\begin{figure*}[t]
\centering
\vspace*{-4pt}
\includegraphics[width=0.955\textwidth,trim={0 135pt 0 15pt},clip]{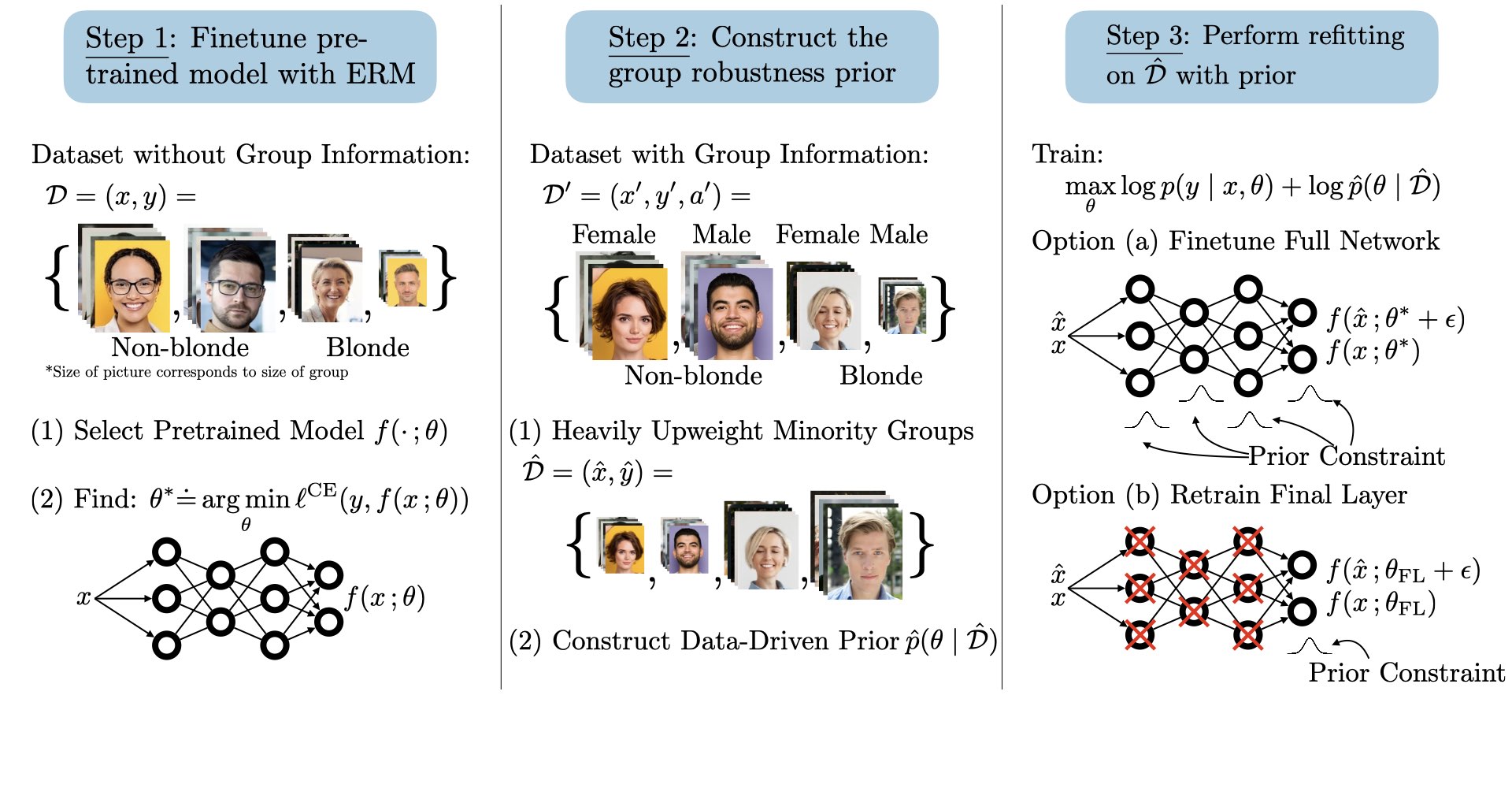}
\vspace*{-3pt}
\caption{
    {\bf Training with a Group-Aware Prior (GAP).}
    Step 1: Train a neural network with empirical risk minimization (ERM).
    Step 2: Construct a group-aware prior by defining a tractable joint distribution that places high probability density on parameters that achieve high worst group accuracy.
    Step 3: Find the most likely parameters under the posterior induced by the group-aware prior and the data.$^{2}$} %
    \label{fig:schematic}
\vspace*{-10pt}
\end{figure*}

\section{INTRODUCTION}
\label{sec:intro}
\vspace{-3pt}

Distribution shifts, frequently occurring in real-world data, have long plagued machine learning models \citep{Quinonero-Candela_2009}.
Empirical risk minimization \citep[ERM; ][]{vapnik}---the minimization of average training loss---is known to generalize poorly under distribution shifts. 
In particular, \emph{subpopulation shifts}---due to attribute and class biases---can cause significantly increased test error on certain population groups, even if the average test error remains low \citep{pmlr-v80-hashimoto18a}.
In many applications, high accuracy on certain subpopulations/groups is essential, and failure to generalize under subpopulation shifts can have severe and harmful consequences \citep{bigdatadisparity, Dastin2018}.

In this paper, we focus on achieving \emph{group robustness}, increasing the worst-case test performance across groups represented in the data, which is crucial for building equitable and effective machine learning systems.
A variety of approaches exist to tackle this problem, including methods that explicitly optimize for worst-case group performance \citep{Sagawa2020Distributionally}, approaches that identify neural network features that lead to improved group robustness \citep{kirichenko2022dfr}, and several techniques that rely on training helper models used to identify shifts and reweighting the data \citep[e.g.,][]{justtraintwice,nam2022SSA}.

Unlike previous work, we approach group robustness from a Bayesian perspective and present a general approach to designing data-driven priors that favor models with high group robustness.
Under such priors, performing Bayesian inference will lead to posterior distributions over neural network parameters that allow the model to fit the training data while also respecting the soft constraints imposed by the prior distribution.

To demonstrate the usefulness of such priors, we construct an example of a simple data-driven {\em group-aware prior} (GAP) distribution over the parameters of a neural network designed to place high probability density on parameter values that induce predictive models that generalize well under subpopulation shifts.
While exact Bayesian inference with non-standard priors can be challenging, we build on the approach presented in \citet{Rudner2023functionspace} to find the most likely parameters under the posterior distribution implied by the prior and the data.
We illustrate the process of training a model with a group-aware prior in \Cref{fig:schematic}.

Data-driven group-aware prior distributions allow for probabilistically principled learning, are modular in their ability to be applied to any likelihood function, and are simple to implement.
In our empirical evaluation, we consider the realistic setting where only a small set of data with group information is available and construct a simple example of a group-aware prior.
We then show that for established subpopulation shift benchmarking tasks \begin{enumerate*}[label=(\roman*)] \item finetuning a previously trained model with a group-aware prior \emph{leads to state-of-the-art results on all benchmarks} and \item only retraining the final layer of a previously trained model with a group-aware prior leads to state-of-the-art results on two of the benchmarks, and remains competitive on the other.\footnotemark\end{enumerate*}
Excitingly, this probabilistic formulation of group robustness opens up new routes for bringing to bear the vast arsenal of Bayesian inference methods to obtain even higher levels of group robustness.

To summarize, our key contributions are as follows:\vspace*{-10pt}
\begin{enumerate}[leftmargin=15pt]
\setlength\itemsep{0pt}
    \item
    We present a general framework for constructing tractable data-driven priors to achieve group robustness under subpopulation shifts. 
    \item
    We design a simple group-aware prior (GAP) that places high probability density on parameter values that lead to high group robustness. %
    \item
    We show empirically that finetuning a previously trained model with this prior leads to state-of-the-art results on standard benchmarking tasks.
    \item
    We consider a more constrained setting in which we freeze a previously trained network and only retrain the final layer.
    We find that even in this highly constrained setting, retraining only a few hundred parameters with group-aware priors leads to state-of-the-art results.
\end{enumerate}
\vspace*{-5pt}
\footnotetext[1]{Our code is available at \url{https://github.com/timrudner/group-aware-priors}.}

\section{BACKGROUND}
\label{sec:background}

\subsection{Learning as Probabilistic Inference}

\footnotetext[2]{All images in this paper are taken from \url{https://unsplash.com} or \url{https://www.istockphoto.com} under unlimited, perpetual, nonexclusive, worldwide license.}

Consider supervised learning problems with $N$ i.i.d. data realizations \mbox{${\calD = \{x_\calD^{(n)}, y_\calD^{(n)}\}_{n=1}^N} = (x_\calD, y_\calD)$} of inputs \mbox{$X \in \calX$} and labels \mbox{$Y \in \calY$} with input space \mbox{$\calX$} and label space \mbox{$\calY$}.
For supervised learning tasks, we define a parametric observation model $p_{Y | X, \Theta}(y \vbar x, \theta; f)$ with a neural network mapping, \mbox{$f(\cdot \,; \theta)$}, and a \textit{prior} distribution over the parameters, $p_{\Theta},(\theta)$ with the goal of inferring a {\em posterior distribution} from the data.

Since, by Bayes' Theorem, the posterior under this model is proportional to the joint probability density given by the product of the likelihood of the parameters under the data $p_{Y | X, \Theta}(y_{\calD} \vbar x_{\calD}, \theta)$ and the prior $p_{\Theta}(\theta)$,\vspace*{-5pt}
\begin{align*}
    p_{\Theta | Y, X}(\theta \vbar y_{\calD}, x_{\calD}) 
    \propto
    p_{Y | X, \Theta}(y_{\calD} \vbar x_{\calD} , \theta) p_{\Theta}(\theta) ,
\end{align*}\\[-15pt]
the most likely parameters under the posterior are given by the mode of $p_{Y | X, \Theta}(y_{\calD} \vbar x_{\calD}, \theta) p_{\Theta}(\theta)$.
Maximum a posteriori (\map) estimation seeks to find this mode, $\theta^{\map}$~\citep{Bishop2006PatternRA,murphy2013probabilistic}.
Under a likelihood that factorizes across the data points given parameters $\theta$,
the MAP optimization objective can be expressed as\vspace*{-5pt}
\begin{align*}
    \calF^{\map}(\theta)
    =
    \sum\nolimits_{n=1}^N \log{p_{Y | X, \Theta}(y^{(n)}_{\calD} \vbar x^{(n)}_{\calD} , \theta)} + \log{p_{\Theta}(\theta)} .
\end{align*}\\[-14pt]
Under Gaussian and categorical likelihood functions, the log-likelihood term in the MAP optimization objective corresponds to a scaled negative mean squared error (MSE) loss function and a negative cross-entropy loss function, respectively.
Similarly, under Gaussian and Laplace priors, the log-density of the prior is proportional to $L_2$ regularization and $L_1$ regularization, respectively.
We will use this probabilistic perspective to obtain a tractable optimization objective that allows incorporating data-driven priors designed to improve the group robustness of neural networks into training.

\subsection{Subpopulation Shifts}

We consider classification problems on data $(x, y) \in \mathcal{X} \times \mathcal{Y}$, where we assume that the data consists of several groups (subpopulations)  $g \in \mathcal{G}$,
which are often defined by a combination of a label $y \in \mathcal{Y}$ and spurious attribute $a \in \mathcal{A}$ (sometimes called {\em environment}).
The attribute $a \in \mathcal{A}$ may or may not be available during training. For instance, in CelebA hair color prediction \citep{celeba}, the labels are `blond' and `brown', and the groups are `non-blond women' ($g_1$), `blond women' ($g_2$), `non-blond men' ($g_3$), and `blond men' ($g_4$) with proportions $44\%$, $14\%$, $41\%$, and $1\%$ of the data, respectively; the group $g_4$ is the minority group, and gender serves as the attribute (spurious feature).

We assume that the training data comes from a mixture of group-wise distributions $p_{\textrm{train}}=\sum_{g \in \mathcal{G}} \alpha_g p_g$, where $\alpha \in \Delta_{|\mathcal{G}|}$ (group weights summing to $1$) and $p_g$ are distributions over elements of the group $g \in \mathcal{G}$.
We speak of {\em subpopulation shift} when the test data comes from a differently weighted distribution $p_{\textrm{test}}=\sum_{g \in \mathcal{G}} \beta_g p_g$, where the weighting $\beta \in \Delta_{|\mathcal{G}|}$ is not known at train-time. 
This leads to the natural goal of maximizing {\em worst group test accuracy} (WGA), that is, the lowest accuracy across groups $\mathcal{G}$ represented in the test data.
We follow the shift taxonomy proposed in \citet{yang2023change}, and our benchmarking datasets exhibit combinations of all of the following three shifts.

{\em Spurious correlations} are present when a label $y$ is correlated with an attribute $a$ in the training distribution but not in the test distribution.
For instance, it was shown that thoracic X-ray images contain spurious correlations between labels and attributes such as patient age, scanning position, or text font, which may not be present in the test data~\citep{heaven21covid,DeGrave2021covid}.

An {\em attribute shift} is present when the distribution of an attribute $a$ for a specific label $y$ and context combination differs between the training and test distributions.
For instance, MultiNLI has a high prevalence of negation words (`no', `never')---one of the attributes.
 
A {\em Class shift} is present when the distribution of classes differs between the training and test distributions.
For example, in the CelebA dataset, the `blond' class in the training set constitutes $15\%$ of the training examples but a larger fraction in the test set.

\section{RELATED WORK}
\label{sec:related_work}

Subpopulation shifts are omnipresent in real-world applications, which has led to a large body of literature concerned with reducing the adverse effects of spurious correlations, attribute shifts, and class shifts on model performance through more robust models.

Ensuring robustness to subpopulation shifts is a longstanding challenge in machine learning.
A broad range of methods has been developed to mitigate different types of subpopulation shifts, including \emph{data reweighting} \citep{pmlr-v177-idrissi22a}, \emph{data augmentation} \citep{zhang2018mixup}, \emph{domain-invariant feature learning} \citep{arjovsky2020invariant, li2018mmd}, and \emph{\mbox{(sub-)group} robustness} methods \citep{Sagawa2020Distributionally, justtraintwice, nam2022SSA, zhang22z,kirichenko2022dfr}.
Several methods designed to achieve high worst group accuracy build on the distributionally robust optimization (DRO) framework \citep{Rahimian_2022}, where worst-case accuracy---instead of average case accuracy---is explicitly maximized during training~\citep{ben2013robust,hu2018does,oren2019distributionally,zhang2020coping}.
Notably, GroupDRO \citep{Sagawa2020Distributionally} has become a standard baseline for group robustness.

{\bf Group Labeling Models.}~
In most real-world settings, obtaining group information is expensive and only feasible for a small number of data points.
To emulate real-world settings where only a small number of training data with group information is available, existing methods have used the validation sets of benchmarking datasets to achieve high worst group accuracy.
\citet{nam2022SSA} and \citet{sohoni2022barack} train group labeling models on group-labeled validation data, generate group labels for the training dataset, and then train a second model on the full dataset using the generated group labels.
In contrast, our proposed method does not require training a group labeling model and instead only requires reweighting the validation set during group-robustness finetuning to obtain state-of-the-art performance, outperforming all group labeling techniques.

{\bf Last-Layer Retraining.}~
A particularly simple and efficient approach to improving group robustness in settings with limited group label availability is {\em deep feature reweighting} \citep[DFR; ][]{kirichenko2022dfr}, which first finetunes a neural network using expected risk minimization (ERM) on a training dataset without group information and then retrains the last layer on a group-balanced reweighting dataset using ERM regularized with an $L_1$-norm between the current and the previously fine-tuned last-layer parameters (see also \citet{izmailov2022feature}, \citet{qiu2023afr}, and \citet{le2023layer} for follow-up works).
This work is notable for its marked simplicity and low complexity compared to related methods, as it only requires last-layer retraining on a validation set to reach state-of-the-art performance on some benchmarks and competitive performance on other benchmarks.
We present results for our method in the setting where we only retrain the final layer with a validation set and show that it outperforms DFR on a majority of benchmarking tasks.
While \citet{kirichenko2022dfr} showed that DFR leads to worse performance when group-robustness finetuning on the full network, we find that group-robustness finetuning the full network with our method leads to significant improvements over last-layer retraining, resulting in state-of-the-art results.

{\bf Data-driven Priors.}~
Prior distributions encode prior information about random variables, but designing informative prior distributions over neural network parameters is challenging in practice due to the limited interpretability of neural networks.
To address this challenge, previous work has proposed data-driven priors by pertaining neural networks and using the trained parameters to specify priors for related downstream tasks.
For example, \citet{shwartz-ziv2022pretrain} propose to reshape the loss surface using data-driven priors.
These informative priors are learned from the source task---similar to the pre-training paradigm---and lead to improved transfer generalization.
Taking an alternative route, \citet{Rudner2023functionspace} propose {\em function-space empirical Bayes}, deriving an optimization objective that allows approximating the most likely parameters under posteriors (i.e., the maximum a posteriori estimate) computed from sophisticated data-driven priors and use it to learn models with improved uncertainty quantification.
We use the optimization objective proposed in \citet{Rudner2023functionspace} to perform maximum a posteriori estimation with group-aware priors.

{\bf Generalization and Parameter Perturbations.}
Prior work on improving generalization of neural networks, notably {\em Sharpness-Aware Minimization} \citep[SAM; ][]{foret2021sharpnessaware} and {\em Stochastic Weight-Averaging} \citep[SWA; ][]{izmailov2018swa}, has attempted to improve model performance by finding flatter minima of the loss landscape.
The underlying idea is that minimizing a loss under different types of perturbations to the neural network parameters will steer the learned neural network parameters into regions of parameter space where small perturbations to the parameters do not lead to significant increases in the training loss and ultimately lead to learned parameters that correspond to flat loss minima, which in turn have been related to improved generalization.

\section{GROUP-AWARE PRIORS}
\label{sec:method}

In this section, we first define a general family of group-aware priors.
Then, we use this framework to develop a simple and scalable instantiation of a group-aware prior.
Finally, we describe how to apply the maximum a posteriori estimation procedure proposed in \citet{Rudner2023functionspace} to use this prior to train a neural network.

\subsection{A Family of Group-Aware Priors}
\label{sec:group_robustness_prior_family}

We begin by specifying the auxiliary inference problem.
Let $\hat{x}$ be a set of \emph{context points} \mbox{$\hat{x} = \{ \hat{x}_{1}, ..., \hat{x}_{M} \}$} with corresponding \emph{context labels} $\hat{y} = \{ \hat{y}_{1}, ..., \hat{y}_{M} \}$, and let $\hat{Z}$ be a Bernoulli random variable denoting whether a given set of neural network parameters induces predictions with some desired property (e.g., high uncertainty on certain evaluation points, high accuracy on evaluation points with a certain group attribute, etc.).
We define an auxiliary likelihood function $\hat{p}_{\smash{\hat{Z} \vbar \Theta}}(\hat{z} \vbar \theta ; f, p_{\smash{\hat{X}, \hat{Y}}})$---which denotes the likelihood of observing a yet-to-be-specified outcome $\hat{z}$ under $\hat{p}_{\smash{\hat{Z} \vbar \Theta}}$ given $\theta$ and $p_{\smash{\hat{X}, \hat{Y}}}$---and a prior over the model parameters, $p_{\Theta}(\theta)$.
For notational simplicity, we will drop the subscripts going forward except when needed for clarity.
By Bayes' Theorem, the posterior under this model and observation $\hat{z}$ is given by
\begin{align}
    \hat{p}(\theta \vbar \hat{z} ; f, p_{\smash{\hat{X}, \hat{Y}}})
    =
    \frac{ \hat{p}(\hat{z} \vbar \theta ; f, p_{\smash{\hat{X}, \hat{Y}}}) p(\theta) }{ \hat{p}(\hat{z} \vbar \theta ; f, p_{\smash{\hat{X}, \hat{Y}}}) } .
\end{align}
To define a family of data-driven priors that place high probability density on neural network parameter values that induce predictive functions that achieve high group robustness, we define a specific Bernoulli auxiliary observation model $\hat{p}_{\smash{\hat{Z} \vbar \Theta}}$ in which $\hat{Z} = 1$ denotes the outcome of `achieving group robustness' and $\hat{p}(\hat{z} = 1 \vbar \theta ; f, p_{\smash{\hat{X}, \hat{Y}}})$ denotes the likelihood of $\hat{z} = 1$ given $\theta$ and $p_{\smash{\hat{X}, \hat{Y}}}$.
We can now define a general family of group-aware priors by specifying the Bernoulli observation model
\begin{align}
\begin{split}
    \hat{p}(\hat{z} = 1 \vbar \theta ; f, p_{\smash{\hat{X}, \hat{Y}}})
    &
    =
    \exp(- \lambda \mathbb{E}_{p_{\smash{\hat{X}, \hat{Y}}}}[c(\hat{X}, \hat{Y}, \theta) ] )
    \\
    \hat{p}(\hat{z} = 0 \vbar \theta ; f, p_{\smash{\hat{X}, \hat{Y}}})
    &
    =
    1 - \hat{p}(\hat{z} = 1 \vbar \theta ; f, p_{\smash{\hat{X}, \hat{Y}}}) ,
\end{split}
\end{align}
where $c : \mathcal{X} \times \mathcal{Y} \times \mathbb{R}^{P} \rightarrow \mathbb{R}_{\geq}$ is a `cost' function and $\lambda > 0$ is a scaling parameter.
By specifying an auxiliary dataset with $\hat{\calD} = \hat{z} = \{ 1, ..., 1\}$ and a distribution $p_{\smash{\hat{X}, \hat{Y}}}$ we obtain a posterior $\hat{p}(\theta \vbar \hat{z} ; f, p_{\smash{\hat{X}, \hat{Y}}})$, the distribution over neural network parameters that we \emph{would} infer if we observed outcomes $\hat{z} = \{ 1, ..., 1\}$ under the likelihood defined above.
As with every Bayesian method, the quality of this posterior is determined by the quality of the observation model $\hat{p}_{\smash{\hat{Z} \vbar \Theta}}$, the data, and the prior.
Therefore, if the observation model is poor, so will be the posterior.
As a result, the main challenge in designing useful group-aware priors is to construct an observation model---that is, a cost function $c$---that is as well-specified as possible.
The better specified the observation model, the more useful the data-driven prior will be.
Below, we present a specific instantiation of a group-aware prior by proposing a simple cost function, paired with a suitable set of context points and context labels.

\subsection{A Simple Group-Aware Prior}
\label{sec:group_robustness_prior_simple}

To specify a practically useful group-aware prior, we need two ingredients: \begin{enumerate*}[label=(\roman*)] \item We need to specify the distribution $p_{\smash{\hat{X}, \hat{Y}}}$ for which observing $\hat{z} = \{1, ..., 1\}$ would be most informative, and \item we need to specify a cost function $c$ which---for suitably chosen $p_{\smash{\hat{X}, \hat{Y}}}$---is a good proxy for group robustness on unknown test points.\end{enumerate*}
Both (i) and (ii) are generally challenging and could be tackled with sophisticated methods---for example, by learning tailored generative models or handcrafting complex cost functions.
However, to demonstrate the usefulness of group-aware priors, we instead limit ourselves to fixed, prespecified distribution $p_{\smash{\hat{X}, \hat{Y}}}$ and a very simple cost function.

First, to specify a useful context distribution $p_{\smash{\hat{X}, \hat{Y}}}$, we assume that we have access to at least a small dataset, $D_{\textrm{val}}$, with group information and simply upsample the dataset to create a distribution
\begin{align}
    p_{\smash{\hat{X}, \hat{Y}}} = \sum\nolimits_{g \in \mathcal{G}} \alpha_g p_g
    \label{eq:sampling_distribution}
\end{align}
with $\alpha_g = \tilde{\alpha}_g / \sum_{g \in \mathcal{G}} \tilde{\alpha}_g$ and $\tilde{\alpha}_g = (|\calD_{\textrm{val}}| / |\calD_{\textrm{val}}^{g}|)^{\gamma}$, where $|\calD_{\textrm{val}}|$ is the number of data points in $\calD_{\textrm{val}}$, $|\calD_{\textrm{val}}^{g}|$ is the number of data points from group $g$ in $\calD_{\textrm{val}}$, and $\gamma \geq 1$ is a scaling parameter.
The larger the hyperparameter $\gamma$, the stronger rare groups will be upweighted.
The smaller $|\calD^{g}_{\textrm{val}}|$  as a fraction of $|\calD_{\textrm{val}}|$, the larger $\alpha_g$ will be.

Second, to specify a useful cost function, we define
\begin{align}
    c(\hat{x}, \hat{y}, \theta)
    \defines
    \ell(\hat{y}, f(\hat{x} ; \theta + \rho \epsilon(\theta))) 
\end{align}
where $\rho > 0$ is a scaling parameter and $\ell$ is a cross-entropy loss function for classification tasks and a mean-squared error loss function for regression tasks.
$\epsilon(\theta)$ is a worst-case perturbation to the model parameters proposed in \citet{foret2021sharpnessaware} and given by
\begin{align}
    \epsilon(\theta, \hat{x}, \hat{y})
    \defines
    \perp \frac{\nabla_{\theta} \ell(\hat{y}, f(\hat{x} ; \theta)) }{ \| \nabla_{\theta} \ell(\hat{y}, f(\hat{x} ; \theta)) \|_{2}} ,
\end{align}
where $\perp$ is the \texttt{stop\_gradient} operator.
The intiution for this cost function is simple: we know that achieving low loss tends to correspond to good generalization, but given that the context dataset has---by assumption---few data points from rare groups, generalizing to test points from these groups is difficult.
To design a cost function that leads to a data-driven prior with high probability density on parameters that enable good generalization to points from rare groups, we add a worst-case perturbation defined to lead to a maximum increase in the loss, as proposed in \citet{foret2021sharpnessaware}.

Unfortunately, the full posterior $\hat{p}(\theta \vbar \hat{z} ; f, p_{\smash{\hat{X}, \hat{Y}}})$ is intractable, since the parameters $\theta$ appear non-linearly in $c$.
However, taking the log of the analytically tractable joint density $\hat{p}(\hat{z} \vbar \theta ; f, p_{\smash{\hat{X}, \hat{Y}}}) p(\theta)$, we obtain\vspace*{-3pt}
\begin{align}
\begin{split}
    &
    \log \hat{p}(\hat{z} \vbar \theta ; f, p_{\smash{\hat{X}, \hat{Y}}}) + \log p(\theta)
    \\
    &
    ~~~
    \propto
    - \mathbb{E}_{p_{\smash{\hat{X}, \hat{Y}}}} [ \ell(\hat{Y}, f(\hat{X} ; \theta + \rho \epsilon(\theta))) ] + \log p(\theta)
    \\
    &
    ~~~\,
    \defines
    \mathcal{J}(\theta, p_{\smash{\hat{X}, \hat{Y}}}) ,
\end{split}
\end{align}
with proportionality up to an additive constant independent of $\theta$.
This implies that
\vspace*{-4pt}
\begin{align*}
    \argmax\nolimits_{\theta} \hat{p}(\theta \vbar \hat{z} ; f, p_{\smash{\hat{X}, \hat{Y}}})
    =
    \argmax\nolimits_{\theta} \mathcal{J}(\theta, p_{\smash{\hat{X}, \hat{Y}}}) ,
\end{align*}\\[-15pt]
that is, maximizing the analytically tractable expression $\mathcal{J}(\theta, p_{\smash{\hat{X}, \hat{Y}}})$ with respect to $\theta$ is mathematically equivalent to maximizing the posterior $\hat{p}(\theta \vbar \hat{z} ; f , p_{\smash{\hat{X}, \hat{Y}}})$.
Next, we will show how to use this insight to train neural networks with this prior.

\vspace*{-4pt}
\subsection{Maximum A Posteriori Estimation}
\label{sec:ebvi}

To train a neural network with data-driven group-aware priors, we follow the approach described in \citet{Rudner2023functionspace} and derive a tractable objective function for neural network training with group-aware priors that alows us to find the most likely neural network parameters under the posterior distribution,
\begin{align}
\begin{split}    
\label{eq:posterior_gap}
    p(\theta \vbar y_{\calD}, x_{\calD}, \hat{z} ; f , p_{\smash{\hat{X}, \hat{Y}}})
    =
    \frac{
    p(y_{\calD} \vbar x_{\calD} , \theta) \hat{p}(\theta \vbar \hat{z} ; f , p_{\smash{\hat{X}, \hat{Y}}})}{p(y_{\calD} \vbar x_{\calD}, \hat{z} ; f , p_{\smash{\hat{X}, \hat{Y}}})} .
\end{split}
\end{align}\\[-10pt]
To do so, we use two insights: (i) that maximum a posteriori estimation only requires an analytically tractable function proportional to the log-joint distribution $\log ( p(y_{\calD} \vbar x_{\calD} , \theta) \hat{p}(\theta \vbar \hat{z} ; f , p_{\smash{\hat{X}, \hat{Y}}}) )$ and (ii) that that the group-aware prior $\hat{p}(\theta \vbar \hat{z} ; f , p_{\smash{\hat{X}, \hat{Y}}})$---which is itself an analytically intractable posterior distribution---is proportional to the analytically tractable function $\mathcal{J}(\theta, p_{\smash{\hat{X}, \hat{Y}}})$.
Noting that by taking the logarithm of the posterior in \Cref{eq:posterior_gap}, we get
\begin{align}
\begin{split}
    &
    \log p(\theta \vbar y_{\calD}, x_{\calD} ; f , p_{\smash{\hat{X}, \hat{Y}}})
    \\
    &
    ~~~
    \propto
    \log p(y_{\calD} \vbar x_{\calD} , \theta) + \log \hat{p}(\theta \vbar \hat{z} ; f , p_{\smash{\hat{X}, \hat{Y}}})
    \\
    &
    ~~~
    \propto
    \log p(y_{\calD} \vbar x_{\calD} , \theta) + \mathcal{J}(\theta, p_{\smash{\hat{X}, \hat{Y}}}) ,
\end{split}
\end{align}
which we can write in the form of a standard optimization objective
\begin{align*}
    \calF(\theta)
    \defines
    \sum\nolimits_{n=1}^N
    \hspace*{-1pt}
    \log p(y^{(n)}_{\calD} \vbar x^{(n)}_{\calD} , \theta)
    \hspace*{-1pt}
    +
    \hspace*{-1pt}
    \mathcal{J}(\theta, p_{\smash{\hat{X}, \hat{Y}}})
    ,
\end{align*}%
where $\log p(y^{(n)}_{\calD} \vbar x^{(n)}_{\calD} , \theta)$ is a data-fit term and $\mathcal{J}(\theta, p_{\smash{\hat{X}, \hat{Y}}})$ is a regularization term that favors parameter values that have a high level of group robustness.
For a Gaussian prior $p(\theta) = \calN(\theta ; \mu, \tau_{\theta}^{-1})$, we have \mbox{$\log p(\theta) \propto -\frac{\tau_{\theta}}{2} \|\theta - \mu\|_{2}^{2} $}, where $\mu$ could be any set of prior mean parameters, the final minimization objective takes the simple form
\begin{align}
\begin{split}
\label{eq:e-map-objective}
    \calL(\theta)
    &
    =
    \sum\nolimits_{n=1}^N
    \hspace*{-1pt}
    \ell(y^{(n)}_{\calD}, f(x^{(n)}_{\calD} ; \theta))
    +
    \frac{\tau_{\theta}}{2} \|\theta - \mu \|_{2}^{2}
    \\
    &
    \quad~~
    +
    \lambda \mathbb{E}_{p_{\smash{\hat{X}, \hat{Y}}}} [ \ell(\hat{Y}, f(\hat{X} ; \theta + \rho \epsilon(\theta))) ] ,
\end{split}
\end{align}%
which we can compute via simple Monte Carlo estimation as
\begin{align}
\begin{split}
\label{eq:e-map-objective-mc}
    \hspace*{-4pt}\hat{\calL}(\theta)
    &
    \defines
    \underbrace{\sum\nolimits_{n=1}^N
    \hspace*{-1pt}
    \ell(y^{(n)}_{\calD}, f(x^{(n)}_{\calD} ; \theta))
    +
    \frac{\tau_{\theta}}{2} \|\theta - \mu \|_{2}^{2}}_{\textrm{standard $L_{2}$-regularized loss}}
    \\
    &
    \quad~~
    +
    \underbrace{\frac{\lambda}{S} \sum\nolimits_{s=1}^S \ell(\hat{y}^{(s)}, f(\hat{x}^{(s)} ; \theta + \rho \epsilon(\theta)))}_{\textrm{robustness regularization}} ,
\end{split}
\end{align}\\[-10pt]
where $(\hat{x}^{(s)}, \hat{y}^{(s)}) \sim p_{\smash{\hat{X}, \hat{Y}}}$.
This objective is amenable to optimization with stochastic gradient descent.

\subsection{Practical Considerations}

\textbf{Deconstructing Loss Components.}~
The optimization objective of \Cref{eq:e-map-objective-mc} contains two distinct terms: (i) The {\em standard $L_2$-regularized loss} is computed on the training set and does not require any group labels. The {\em robustness regularization} is evaluated on the context distribution $p_{\smash{\hat{X}, \hat{Y}}}$ sampled from the validation set.
Defining $p_{\smash{\hat{X}, \hat{Y}}}$ requires group labels to upweight the data as shown in \Cref{eq:sampling_distribution}.
Lastly, it is worth noting that only the robustness regularization incorporates a perturbation of the parameters.
This reflects that we would like the prior to favor flatter minima that specifically improve generalization to minority groups.

\textbf{Hyperparameters.}~
The optimization objective in \Cref{eq:e-map-objective-mc} has three additional hyperparameters, compared to training with $L_2$-regularized ERM.
The parameter $\lambda$ governs the strength of the empirical prior, the parameter $\rho$ governs the strength of the parameter perturbation, and the parameter $\gamma$ governs how strongly the distribution is reweighted.

\textbf{Computational Complexity.}~
The objective requires two additional forward passes on the $M$ points sampled from the context distribution: one to compute epsilon and one to compute the loss under the parameter perturbation.
In practice, this slowdown in training speed is not a problem since we only ever train for a handful of epochs, as detailed in \Cref{appsec:experiment}.

\begin{table*}[t]
\centering
\small
\caption{
    Worst group and average accuracy on the test set of our method against a variety of other baselines in recent literature. We follow \citet{Sagawa2020Distributionally} and reweigh the test accuracy for each group based on their proportion in the training data.
    The Group Info column details whether a method uses group labels in the training and validation dataset and whether it uses an auxiliary group labeling model.
    Accuracies of our method are estimated over ten trials.
    For a description of the baselines methods, see \Cref{sec:baselines}.
    We report the standard error of the mean, which sometimes requires adjusting the error bars of other baselines.
    The best-performing method is highlighted in gray and bolded, and the second-best-performing method is only bolded.
}
\label{tab:results}
\begin{tblr}{width=\textwidth,
  colspec={X[2.5]X[0.5]X[0.5]X[0.5]X[1.5]X[1.5]X[1.5]X[1.5]X[1.5]X[1.5]},
  row{1} = {}{font=\bfseries},
  column{2,3,4} = {}{c},
  cell{1}{1} = {r=2}{},
  cell{1}{2} = {c=3}{c},
  cell{1}{5} = {c=2}{c},
  cell{1}{7} = {c=2}{c},
  cell{1}{9} = {c=2}{c},
  cell{-}{3,4} = {}{c},
  cell{2-13}{5,7,9} = {}{c},
  cell{2-13}{6,8,10} = {}{c},
  cell{11}{5,6,9,10} = {}{bg=lightgray!30, font=\boldmath},
  cell{10}{7} = {}{bg=lightgray!30, font=\boldmath},
  cell{11}{7} = {}{bg=lightgray!30, font=\boldmath},
  cell{9}{8} = {}{bg=lightgray!30, font=\boldmath},
  cell{9}{9} = {}{font=\boldmath},
  cell{7}{10} = {}{font=\boldmath},
  cell{6}{8} = {}{font=\boldmath},
  cell{10}{5,6} = {}{font=\boldmath},
  hline{3,4,10} = {-}{},
  hline{2} = {5-11}{},
}
\toprule
Method & Group Info & & & Waterbirds&              & CelebA &              & MultiNLI &            \\
& Tr. & Val. & Aux. & Worst              & Average      & Worst             & Average      & Worst           & Average    \\
ERM            &  N & N & N & $74.9\pms{1.0}$ & $98.1\pms{0.0}$  & $46.9\pms{1.3}$& $95.3\pms{0.0}$ & $65.9\pms{0.1}$  & $82.8\pms{0.0}$      \\
JTT            &  N & Y & Y & $86.7$             & $93.3$     & $81.1$          & $88.0$  & $72.6$            & $78.6$            \\
CnC            &  N & Y & Y & $88.5\pms{0.2}$    & $90.9\pms{0.1}$& $88.8\pms{0.5}$ & $89.9\pms{0.3}$ & ---             & ---              \\
SSA            &  N & Y & Y & $89.0\pms{0.3}$    & $92.2\pms{0.5}$& $89.8\pms{0.8}$    & $92.8\pms{0.1}$ & $76.6\pms{0.4}$ & $79.9\pms{0.5}$      \\
DFR            &  N & Y & N & $92.9\pms{0.1}$    & $94.2\pms{0.2}$ & $88.3\pms{0.5}$& $91.3\pms{0.1}$ & $74.7\pms{0.3}$ & $82.1\pms{0.1}$      \\
SUBG           &  Y & Y & N & $89.1\pms{0.5}$    & ---          & $85.6\pms{1.0}$ & ---   & $68.9\pms{0.4}$  &  ---                \\
G-DRO          &  Y & Y & N & $91.4$    & $93.5$  & $88.9$     & $92.9$   & $77.7$   & $81.4$      \\
{\bf GAP} {\small Last Layer}  &  N & Y & N & $93.2\pms{0.2}$ & $94.6\pms{0.2}$ & $90.2\pms{0.3}$ & $91.7\pms{0.2}$ & $74.3\pms{0.2}$ & $81.9\pms{0.0}$   \\
{\bf GAP} {\small All Layers}&  N & Y & N & $93.8\pms{0.1}$ & $95.6\pms{0.1}$ & $90.2\pms{0.3}$ & $91.5\pms{0.1}$ & $\mathbf{77.8\pms{0.6}}$ & $82.5\pms{0.1}$   \\
\bottomrule
\end{tblr}
\vspace*{3pt}
\end{table*}

\section{EMPIRICAL EVALUATION}\label{sec:emp_eval}

Our experimental evaluation has two moving pieces: (i) datasets and (ii) last-layer retraining versus full finetuning on the group-labeled validation set. We outline the datasets we use, prior benchmark we compare to (including current state-of-the-art), and our experimental setup and show that our method achieves new SOTA or is competitive with the best methods. We also discuss the components that go into our designed prior and provide ablation studies to show their impact.

\subsection{Datasets}

We evaluate our method on both image classification and text datasets that are commonly used to benchmark
the performance of group robustness methods.

\textbf{Waterbirds.}~
Waterbirds is a binary image classification problem generated synthetically by combining images of birds from the CUB dataset \citep{waterbirds1} and backgrounds from the Places dataset \citep{waterbirds2}; the class corresponds to either land- or waterbird.
$73\%$ of images correspond to the majority group (waterbirds on water), $22\%$ are landbirds on land and a sharply pronounced minority group of $1\%$ landbirds on water, as well as $4\%$ waterbirds on land.
The distribution of backgrounds (land/water) on the validation and test sets is balanced.

\textbf{CelebA.}~
We consider a binary classification problem (`blond vs. non-blonde hair color') with gender serving as the spurious feature.
$94\%$ of images with the blond labels show females.
Models were trained on pixel intensities at the top of each image into a binary `blonde vs. not-blonde' label.
No individual face characteristics, landmarks, keypoints, facial mapping, metadata, or any other information was used for training.

\textbf{MultiNLI.}~
MultiNLI is a text classification problem where the task is to classify the relationship between a given pair of sentences as a contradiction, entailment, or neither.
In this dataset, the presence of negation words (e.g., `never') in the second sentence is spuriously correlated with the `contradiction' class.

\subsection{Baselines}
\label{sec:baselines}

We consider seven baseline methods that make different assumptions about the availability of group attributes at training time. Empirical risk minimization (ERM) represents conventional training without any procedures for improving worst group accuracy. 
Just Train Twice \citep[JTT; ][]{justtraintwice} is a method that detects the minority group examples on train data, only using group labels on the validation set to tune hyper-parameters.
Correct-n-Contrast \citep[CnC; ][]{zhang22z} detects the minority group examples similarly to JTT and uses a contrastive objective to learn representations robust to spurious correlations.
Group DRO \citep{Sagawa2020Distributionally} uses group information to train on a worst group loss objective and is a commonly used baseline. 
Deep Feature Reweighting \citep[DFR; ][]{kirichenko2022dfr}, uses a network finetuned with ERM and performs last-layer feature reweighting.
SUBG \citep{pmlr-v177-idrissi22a} is carefully tuned ERM on a random subset of the data where the groups are equally represented.
Finally, Spread Spurious Attribute \citep[SSA; ][]{nam2022SSA} attempts to fully exploit the group-labeled validation data with a semi-supervised approach that propagates the group labels to the training data.

All baselines use pretrained models (a ResNet-50, pretrained on ImageNet1K for Waterbirds and CelebA, and a pretrained BERT model for MultiNLI), and all of them finetune the entire network except for DFR, where finetuning is only performed on the last layer (after full-parameter ERM training).

\vspace*{-2pt}
\subsection{Experimental setup}

For all experiments, we import a pretrained model and finetune all layers of the network using ERM on the target training dataset. 
For vision datasets, we finetune a ResNet-50 \citep{he2016resnet} pretrained on ImageNet \citep{imagenet} for 30 epochs.
For language datasets, we use pretrained BERT \citep{devlin-etal-2019-bert} and finetune for five epochs. 
We assume that the validation dataset $\calD_{\textrm{val}}$ is group-annotated, and we use $85\%$ of it to sample the context distribution $p_{\smash{\hat{X}, \hat{Y}}}$; the remaining $15\%$ are reserved for hyperparameter tuning.
The total size of the validation datasets is 19,867 (out of 202,599) for CelebA, 1,199 (out of 11,788) for Waterbirds, and 82,462 (out of 412,349) for MultiNLI.
In all cases, we heavily upweight the minority groups in the robust regularization term with $\gamma=4, 1.5$, and $2$ for Waterbirds, CelebA, and MultiNLI, respectively.
We use adversarial perturbations of the parameters in the robust regularization term with strength $\rho=0.15$ for Waterbirds and CelebA and $\rho=0.1$ for MultiNLI.
We use the same configurations for last-layer and full-network training.
For further details, see \Cref{appsec:experiment}.

\subsection{Results}

\Cref{tab:results} presents our results.
GAP achieves state-of-the-art performance for all three benchmarking tasks, demonstrating the value of optimizing with even a very simple data-driven prior.
In particular, on Waterbirds and MultiNLI, GAP improves both worst-group and average accuracy even when compared to methods that require training data with group labels or use an auxiliary model to create attribute labels.
For CelebA, GAP improves the worst group accuracy over all baselines, with only marginally lower average accuracy.

Perhaps most remarkably, we achieve state-of-the-art and close-to-state-of-the-art performance even when only retraining the last layer using group-labeled validation data.
In particular, GAP applied to the last layer outperforms DFR, another method that only requires last-layer refitting, on the Waterbirds and CelebA) tasks, and is competitive to DFR on MultiNLI.

\subsection{Ablation Studies}

While the group-aware prior constructed in \Cref{sec:group_robustness_prior_simple} is relatively simple, it adds several additional degrees of freedom.
Most notably, it involves a scaled parameter perturbation, as can be seen in the robust regularizer of \Cref{eq:e-map-objective-mc}, meant to favor flatter minima and better generalization to minority groups.
Furthermore, the expected cost function is computed under a context distribution, which we construct by upweighting minority groups in the data as per \Cref{eq:sampling_distribution}. Below, we show the impact of these design choices in two ablation studies (for further details, see \Cref{appsec:experiment}).

{\bf Ablation on Parameter Perturbation.}~ \Cref{tab:ablation} compares the performance of GAP (last-layer retraining only) with and without the perturbation of the parameters in the robustness regularization term.
As expected, optimizing against a worst-case perturbation in the parameters leads to a small decrease in average accuracy, which is largely compensated for by a significant gain in worst group accuracy. 

\begin{table}[t!]
\centering
\small
\caption{{Ablation on Adversarial Perturbation.}
We ablate the scaling parameter $\rho$ in the GAP robustness regularizer (last-layer retraining only).
GAP uses $\rho=0.15$ for Waterbirds and CelebA and $\rho=0.1$ for MultiNLI.
The $\rho=0$ setting indicates no parameter perturbation.
Means and standard errors are estimated from ten trials.
}
\label{tab:ablation}
\begin{tblr}{width=\columnwidth,
  colspec={X[-1]|XXXXXX},
  colsep=3pt,
  cell{1-4}{1} = {}{c},
  cell{1}{1} = {r=2}{},
  cell{1}{2} = {c=2}{c},
  cell{1}{4} = {c=2}{c},
  cell{1}{6} = {c=2}{c},
  cell{2-10}{2,3,4,5,6,7} = {}{c},
  hline{2,3} = {-}{},
  hline{2} = {3-8}{},
}
\toprule
\textbf{$\rho$}& \textbf{Waterbirds}&                 &  \textbf{CelebA}  &                &  \textbf{MultiNLI}  &        \\
               & Worst              & Average         &  Worst            & Average        &  Worst              & Average \\
\textbf{GAP} & $93.2\pms{0.2}$    & $94.6\pms{0.2}$ &  $90.2\pms{0.3}$  & $91.7\pms{0.2}$&  $74.3\pms{0.2}$    & $81.9\pms{0.0}$ \\
$0$          & $92.7\pms{0.2}$    & $94.8\pms{0.1}$ &  $83.2\pms{0.8}$  & $94.1\pms{0.1}$&  $74.0\pms{0.2}$    & $82.2\pms{0.0}$ \\
\bottomrule
\end{tblr}
\vspace*{-1pt}
\end{table}

{\bf Ablation on Context Distribution.}~
We have designed the context distribution $p_{\smash{\hat{X}, \hat{Y}}}$ to upweight the minority groups (see \Cref{eq:sampling_distribution} and the paragraph following it).
In \Cref{tab:ablation2}, we compare different exponential weighting schemes from a completely group-balanced setting ($\gamma=1$) to very strong upweighting ($\gamma=4$) for Waterbirds (for last-layer retraining).
Stronger upweighting of minority groups beyond the balanced setting is beneficial for improved worst-case accuracy.

\begin{table}[t!]
\centering
\small
\caption{{Ablation on Context Distribution.}
We ablate the upweighting strength in the context distribution $p_{\smash{\hat{X}, \hat{Y}}}$, for exponential upweighting schemes with $\gamma\in\{1,2,4\}$ on the Waterbirds dataset using GAP (last-layer retraining).
Means and standard errors are estimated from three trials for $\gamma \in \{1,2\}$ and ten trials for $\gamma=4$. 
}
\label{tab:ablation2}
\begin{tblr}{
  width=\columnwidth,
  colspec={X[2]|X[1]X[1]},
  cell{1}{1} = {r=2}{},
  cell{1}{2} = {c=2}{},
  cell{1-5}{1,2,3} = {}{c},
  hline{2,3} = {-}{},
}
\toprule
\textbf{Upweight Strength} & \textbf{Waterbirds} &  \\
 & Worst & Average \\
Balanced ($\gamma=1$)&  $90.8\pms{0.6}$    & $95.8\pms{0.4}$   \\
\hspace*{-1.6pt}Moderate ($\gamma=2$)&  $93.1\pms{0.5}$    & $95.1\pms{0.3}$   \\
\hspace*{10.1pt}Strong ($\gamma=4$)&  $\mathbf{93.2}\pms{0.2}$    & $\mathbf{94.6}\pms{0.2}$ \\ 
\bottomrule
\end{tblr}
\vspace*{-3pt}
\end{table}

\section{Discussion}
\label{sec:discussion}

We presented a simple probabilistic framework for learning models that are robust to subpopulation shifts using group-aware prior distributions.
Our empirical evaluation has shown that a probabilistically principled---and yet simple---prior distribution over neural network parameters reflecting group robustness desiderata, is able to achieve state-of-the-art performance on standard benchmarking tasks without the need for any pseudo-labeling routine---even in the highly constrained setting of only retraining the last layer.
This is achieved with minimal computational overhead and implementation complexity since, as we showed in \Cref{eq:e-map-objective}, MAP estimation with the simple group-aware prior can be reduced to adding an additional regularization term to the ERM optimization objective.

{\em Flexibility:}
We have presented a very simple first example of a group-aware prior, which has already proven to be effective at improving robustness to subpopulation shifts.
However, we are not limited to such simple priors.
We have derived a general family of group robustness priors parameterized by a cost function and a context distribution, each of which can be specified using sophisticated models that satisfy group robustness desiderata.
For example, a more sophisticated context distribution $p_{\smash{\hat{X}, \hat{Y}}}$ could be defined by learning a generative model or by using a larger dataset without group information in conjunction with a learned group labeling model.

{\em Complementarity:}
As noted above, methods proposed in related work---such as learned group labeling models, including those that do not require any group labels \citep[e.g., ][]{pezeshki2023discovering}, and data reweighting schemes---complement the framework presented in this paper.

{\em Full Bayesian Inference:}
While we used MAP estimation to find the most likely parameters under the posterior, the probabilistic formulation of learning group robustness via uncertainty-aware priors lends itself to Bayesian inference and, as such, opens up routes for bringing to bear the vast arsenal of Bayesian inference methods.
Inferring full posterior distribution using group-aware priors can improve generalization via Bayesian model averaging~\citet{wilson2020Bayesian} and lead to more reliable uncertainty estimation~\citet{Rudner2023functionspace}.

\section{Acknowledgements}

JK acknowledges support through NSF NRT training grant award 1922658.
AGW acknowledges support through NSF HDR-2118310, CDS\&E-MSS 2134216, CAREER IIS-2145492, I-DISRE 193471.
This work was supported in part through the NYU IT High Performance Computing resources, services, and staff expertise.

\bibliography{references,references_dfr}

\begin{thebibliography}{41}
\providecommand{\natexlab}[1]{#1}
\providecommand{\url}[1]{\texttt{#1}}
\expandafter\ifx\csname urlstyle\endcsname\relax
  \providecommand{\doi}[1]{doi: #1}\else
  \providecommand{\doi}{doi: \begingroup \urlstyle{rm}\Url}\fi

\bibitem[Arjovsky et~al.(2020)Arjovsky, Bottou, Gulrajani, and Lopez-Paz]{arjovsky2020invariant}
Martin Arjovsky, Léon Bottou, Ishaan Gulrajani, and David Lopez-Paz.
\newblock Invariant risk minimization, 2020.

\bibitem[Barocas and Selbst(2016)]{bigdatadisparity}
Solon Barocas and Andrew~D. Selbst.
\newblock Big data's disparate impact.
\newblock \emph{California Law Review}, 104\penalty0 (3):\penalty0 671--732, 2016.
\newblock ISSN 00081221.

\bibitem[Ben-Tal et~al.(2013)Ben-Tal, Den~Hertog, De~Waegenaere, Melenberg, and Rennen]{ben2013robust}
Aharon Ben-Tal, Dick Den~Hertog, Anja De~Waegenaere, Bertrand Melenberg, and Gijs Rennen.
\newblock Robust solutions of optimization problems affected by uncertain probabilities.
\newblock \emph{Management Science}, 59\penalty0 (2):\penalty0 341--357, 2013.

\bibitem[Bishop(2006)]{Bishop2006PatternRA}
Christopher~M. Bishop.
\newblock Pattern recognition and machine learning (information science and statistics).
\newblock 2006.

\bibitem[Dastin(2018)]{Dastin2018}
Jeffrey Dastin.
\newblock Amazon scraps secret ai recruiting tool that showed bias against women.
\newblock \emph{Reuters}, 2018.

\bibitem[DeGrave et~al.(2021)DeGrave, Janizek, and Lee]{DeGrave2021covid}
Alex~J. DeGrave, Joseph~D. Janizek, and Su-In Lee.
\newblock Ai for radiographic covid-19 detection selects shortcuts over signal.
\newblock \emph{Nature Machine Intelligence}, 2021.

\bibitem[Devlin et~al.(2019)Devlin, Chang, Lee, and Toutanova]{devlin-etal-2019-bert}
Jacob Devlin, Ming-Wei Chang, Kenton Lee, and Kristina Toutanova.
\newblock {BERT}: Pre-training of deep bidirectional transformers for language understanding.
\newblock In \emph{Proceedings of the 2019 Conference of the North {A}merican Chapter of the Association for Computational Linguistics: Human Language Technologies, Volume 1 (Long and Short Papers)}, pages 4171--4186, Minneapolis, Minnesota, June 2019. Association for Computational Linguistics.
\newblock \doi{10.18653/v1/N19-1423}.

\bibitem[Foret et~al.(2021)Foret, Kleiner, Mobahi, and Neyshabur]{foret2021sharpnessaware}
Pierre Foret, Ariel Kleiner, Hossein Mobahi, and Behnam Neyshabur.
\newblock Sharpness-aware minimization for efficiently improving generalization.
\newblock In \emph{International Conference on Learning Representations}, 2021.

\bibitem[Hashimoto et~al.(2018)Hashimoto, Srivastava, Namkoong, and Liang]{pmlr-v80-hashimoto18a}
Tatsunori Hashimoto, Megha Srivastava, Hongseok Namkoong, and Percy Liang.
\newblock Fairness without demographics in repeated loss minimization.
\newblock In Jennifer Dy and Andreas Krause, editors, \emph{Proceedings of the 35th International Conference on Machine Learning}, volume~80 of \emph{Proceedings of Machine Learning Research}, pages 1929--1938. PMLR, 10--15 Jul 2018.

\bibitem[He et~al.(2016)He, Zhang, Ren, and Sun]{he2016resnet}
Kaiming He, Xiangyu Zhang, Shaoqing Ren, and Jian Sun.
\newblock Deep residual learning for image recognition.
\newblock In \emph{Proceedings of the IEEE conference on computer vision and pattern recognition}, pages 770--778, 2016.

\bibitem[Heaven(2021)]{heaven21covid}
Will~Douglas Heaven.
\newblock Hundreds of ai tools have been built to catch covid. none of them helped., 2021.

\bibitem[Hu et~al.(2018)Hu, Niu, Sato, and Sugiyama]{hu2018does}
Weihua Hu, Gang Niu, Issei Sato, and Masashi Sugiyama.
\newblock Does distributionally robust supervised learning give robust classifiers?
\newblock In \emph{International Conference on Machine Learning}, pages 2029--2037. PMLR, 2018.

\bibitem[Idrissi et~al.(2022)Idrissi, Arjovsky, Pezeshki, and Lopez-Paz]{pmlr-v177-idrissi22a}
Badr~Youbi Idrissi, Martin Arjovsky, Mohammad Pezeshki, and David Lopez-Paz.
\newblock Simple data balancing achieves competitive worst-group-accuracy.
\newblock In Bernhard Schölkopf, Caroline Uhler, and Kun Zhang, editors, \emph{Proceedings of the First Conference on Causal Learning and Reasoning}, volume 177 of \emph{Proceedings of Machine Learning Research}, pages 336--351. PMLR, 11--13 Apr 2022.

\bibitem[Izmailov et~al.(2018)Izmailov, Podoprikhin, Garipov, Vetrov, and Wilson]{izmailov2018swa}
Pavel Izmailov, Dmitrii Podoprikhin, Timur Garipov, Dmitry Vetrov, and {Andrew Gordon} Wilson.
\newblock Averaging weights leads to wider optima and better generalization.
\newblock In \emph{34th Conference on Uncertainty in Artificial Intelligence 2018, UAI 2018}, 34th Conference on Uncertainty in Artificial Intelligence 2018, UAI 2018, pages 876--885. Association For Uncertainty in Artificial Intelligence (AUAI), 2018.

\bibitem[Izmailov et~al.(2022)Izmailov, Kirichenko, Gruver, and Wilson]{izmailov2022feature}
Pavel Izmailov, Polina Kirichenko, Nate Gruver, and Andrew~G Wilson.
\newblock On feature learning in the presence of spurious correlations.
\newblock \emph{Advances in Neural Information Processing Systems}, 35:\penalty0 38516--38532, 2022.

\bibitem[Kirichenko et~al.(2023)Kirichenko, Izmailov, and Wilson]{kirichenko2022dfr}
Polina Kirichenko, Pavel Izmailov, and Andrew~Gordon Wilson.
\newblock Last layer re-training is sufficient for robustness to spurious correlations.
\newblock In \emph{The Eleventh International Conference on Learning Representations}, 2023.

\bibitem[Le et~al.(2023)Le, Schlötterer, and Seifert]{le2023layer}
Phuong~Quynh Le, Jörg Schlötterer, and Christin Seifert.
\newblock Is last layer re-training truly sufficient for robustness to spurious correlations?, 2023.

\bibitem[Li et~al.(2018)Li, Pan, Wang, and Kot]{li2018mmd}
Haoliang Li, Sinno~Jialin Pan, Shiqi Wang, and Alex~C. Kot.
\newblock Domain generalization with adversarial feature learning.
\newblock In \emph{2018 IEEE/CVF Conference on Computer Vision and Pattern Recognition}, pages 5400--5409, 2018.
\newblock \doi{10.1109/CVPR.2018.00566}.

\bibitem[Liu et~al.(2021)Liu, Haghgoo, Chen, Raghunathan, Koh, Sagawa, Liang, and Finn]{justtraintwice}
Evan~Z Liu, Behzad Haghgoo, Annie~S Chen, Aditi Raghunathan, Pang~Wei Koh, Shiori Sagawa, Percy Liang, and Chelsea Finn.
\newblock Just train twice: Improving group robustness without training group information.
\newblock In Marina Meila and Tong Zhang, editors, \emph{Proceedings of the 38th International Conference on Machine Learning}, volume 139 of \emph{Proceedings of Machine Learning Research}, pages 6781--6792. PMLR, 18--24 Jul 2021.

\bibitem[Liu et~al.(2015)Liu, Luo, Wang, and Tang]{celeba}
Ziwei Liu, Ping Luo, Xiaogang Wang, and Xiaoou Tang.
\newblock Deep learning face attributes in the wild.
\newblock In \emph{Proceedings of International Conference on Computer Vision (ICCV)}, December 2015.

\bibitem[Loshchilov and Hutter(2017)]{loshchilov2017decoupled}
Ilya Loshchilov and Frank Hutter.
\newblock Decoupled weight decay regularization.
\newblock \emph{arXiv preprint arXiv:1711.05101}, 2017.

\bibitem[Murphy(2013)]{murphy2013probabilistic}
Kevin~P. Murphy.
\newblock \emph{Machine learning: a probabilistic perspective}.
\newblock MIT Press, Cambridge, Mass. [u.a.], 2013.
\newblock ISBN 9780262018029 0262018020.

\bibitem[Nam et~al.(2022)Nam, Kim, Lee, and Shin]{nam2022SSA}
Junhyun Nam, Jaehyung Kim, Jaeho Lee, and Jinwoo Shin.
\newblock Spread spurious attribute: Improving worst-group accuracy with spurious attribute estimation.
\newblock In \emph{International Conference on Learning Representations}, 2022.

\bibitem[Oren et~al.(2019)Oren, Sagawa, Hashimoto, and Liang]{oren2019distributionally}
Yonatan Oren, Shiori Sagawa, Tatsunori~B Hashimoto, and Percy Liang.
\newblock Distributionally robust language modeling.
\newblock \emph{arXiv preprint arXiv:1909.02060}, 2019.

\bibitem[Pezeshki et~al.(2023)Pezeshki, Bouchacourt, Ibrahim, Ballas, Vincent, and Lopez-Paz]{pezeshki2023discovering}
Mohammad Pezeshki, Diane Bouchacourt, Mark Ibrahim, Nicolas Ballas, Pascal Vincent, and David Lopez-Paz.
\newblock Discovering environments with xrm, 2023.

\bibitem[Qiu et~al.(2023)Qiu, Potapczynski, Izmailov, and Wilson]{qiu2023afr}
Shikai Qiu, Andres Potapczynski, Pavel Izmailov, and Andrew~Gordon Wilson.
\newblock {Simple and Fast Group Robustness by Automatic Feature Reweighting}.
\newblock \emph{International Conference on Machine Learning (ICML)}, 2023.

\bibitem[Qui\~nonero Candela(2009)]{Quinonero-Candela_2009}
Joaquin Qui\~nonero Candela.
\newblock \emph{Dataset shift in machine learning}.
\newblock MIT Press, 2009.

\bibitem[Rahimian and Mehrotra(2022)]{Rahimian_2022}
Hamed Rahimian and Sanjay Mehrotra.
\newblock Frameworks and results in distributionally robust optimization.
\newblock \emph{Open Journal of Mathematical Optimization}, 3:\penalty0 1--85, jul 2022.
\newblock \doi{10.5802/ojmo.15}.

\bibitem[Rudner et~al.(2023)Rudner, Kapoor, Qiu, and Wilson]{Rudner2023functionspace}
Tim G.~J. Rudner, Sanyam Kapoor, Shikai Qiu, and Andrew~Gordon Wilson.
\newblock Function-space regularization in neural networks: A probabilistic perspective.
\newblock In \emph{Proceedings of the 40th International Conference on Machine Learning}, ICML'23. JMLR.org, 2023.

\bibitem[Russakovsky et~al.(2015)Russakovsky, Deng, Su, Krause, Satheesh, Ma, Huang, Karpathy, Khosla, Bernstein, Berg, and Fei-Fei]{imagenet}
Olga Russakovsky, Jia Deng, Hao Su, Jonathan Krause, Sanjeev Satheesh, Sean Ma, Zhiheng Huang, Andrej Karpathy, Aditya Khosla, Michael Bernstein, Alexander~C. Berg, and Li~Fei-Fei.
\newblock {ImageNet Large Scale Visual Recognition Challenge}.
\newblock \emph{International Journal of Computer Vision (IJCV)}, 115\penalty0 (3):\penalty0 211--252, 2015.
\newblock \doi{10.1007/s11263-015-0816-y}.

\bibitem[Sagawa et~al.(2020)Sagawa, Koh, Hashimoto, and Liang]{Sagawa2020Distributionally}
Shiori Sagawa, Pang~Wei Koh, Tatsunori~B. Hashimoto, and Percy Liang.
\newblock Distributionally robust neural networks.
\newblock In \emph{International Conference on Learning Representations}, 2020.

\bibitem[Shwartz-Ziv et~al.(2022)Shwartz-Ziv, Goldblum, Souri, Kapoor, Zhu, LeCun, and Wilson]{shwartz-ziv2022pretrain}
Ravid Shwartz-Ziv, Micah Goldblum, Hossein Souri, Sanyam Kapoor, Chen Zhu, Yann LeCun, and Andrew~Gordon Wilson.
\newblock Pre-train your loss: Easy bayesian transfer learning with informative priors.
\newblock In Alice~H. Oh, Alekh Agarwal, Danielle Belgrave, and Kyunghyun Cho, editors, \emph{Advances in Neural Information Processing Systems}, 2022.

\bibitem[Sohoni et~al.(2022)Sohoni, Sanjabi, Ballas, Grover, Nie, Firooz, and Re]{sohoni2022barack}
Nimit~Sharad Sohoni, Maziar Sanjabi, Nicolas Ballas, Aditya Grover, Shaoliang Nie, Hamed Firooz, and Christopher Re.
\newblock {BARACK}: Partially supervised group robustness with guarantees.
\newblock In \emph{ICML 2022: Workshop on Spurious Correlations, Invariance and Stability}, 2022.

\bibitem[Vapnik(1998)]{vapnik}
Vladimir Vapnik.
\newblock \emph{Statistical learning theory}.
\newblock Wiley, 1998.
\newblock ISBN 978-0-471-03003-4.

\bibitem[Wah et~al.(2011)Wah, Branson, Welinder, Perona, and Belongie]{waterbirds1}
Catherine Wah, Steve Branson, Peter Welinder, Pietro Perona, and Serge Belongie.
\newblock \emph{The Caltech-UCSD Birds-200-2011 Dataset}.
\newblock Jul 2011.

\bibitem[Wilson and Izmailov(2020)]{wilson2020Bayesian}
Andrew~Gordon Wilson and Pavel Izmailov.
\newblock {B}ayesian deep learning and a probabilistic perspective of generalization.
\newblock In Hugo Larochelle, Marc'Aurelio Ranzato, Raia Hadsell, Maria{-}Florina Balcan, and Hsuan{-}Tien Lin, editors, \emph{Advances in Neural Information Processing Systems 33: Annual Conference on Neural Information Processing Systems 2020, NeurIPS 2020, December 6-12, 2020, virtual}, 2020.

\bibitem[Yang et~al.(2023)Yang, Zhang, Katabi, and Ghassemi]{yang2023change}
Yuzhe Yang, Haoran Zhang, Dina Katabi, and Marzyeh Ghassemi.
\newblock Change is hard: A closer look at subpopulation shift.
\newblock In \emph{International Conference on Machine Learning}, 2023.

\bibitem[Zhang et~al.(2018)Zhang, Cisse, Dauphin, and Lopez-Paz]{zhang2018mixup}
Hongyi Zhang, Moustapha Cisse, Yann~N. Dauphin, and David Lopez-Paz.
\newblock mixup: Beyond empirical risk minimization.
\newblock In \emph{International Conference on Learning Representations}, 2018.

\bibitem[Zhang et~al.(2020)Zhang, Menon, Veit, Bhojanapalli, Kumar, and Sra]{zhang2020coping}
Jingzhao Zhang, Aditya Menon, Andreas Veit, Srinadh Bhojanapalli, Sanjiv Kumar, and Suvrit Sra.
\newblock Coping with label shift via distributionally robust optimisation.
\newblock \emph{arXiv preprint arXiv:2010.12230}, 2020.

\bibitem[Zhang et~al.(2022)Zhang, Sohoni, Zhang, Finn, and Re]{zhang22z}
Michael Zhang, Nimit~S Sohoni, Hongyang~R Zhang, Chelsea Finn, and Christopher Re.
\newblock Correct-n-contrast: a contrastive approach for improving robustness to spurious correlations.
\newblock In Kamalika Chaudhuri, Stefanie Jegelka, Le~Song, Csaba Szepesvari, Gang Niu, and Sivan Sabato, editors, \emph{Proceedings of the 39th International Conference on Machine Learning}, volume 162 of \emph{Proceedings of Machine Learning Research}, pages 26484--26516. PMLR, 17--23 Jul 2022.

\bibitem[Zhou et~al.(2018)Zhou, Lapedriza, Khosla, Oliva, and Torralba]{waterbirds2}
Bolei Zhou, Agata Lapedriza, Aditya Khosla, Aude Oliva, and Antonio Torralba.
\newblock Places: A 10 million image database for scene recognition.
\newblock \emph{IEEE Transactions on Pattern Analysis and Machine Intelligence}, 40\penalty0 (6):\penalty0 1452--1464, 2018.
\newblock \doi{10.1109/TPAMI.2017.2723009}.

\end{thebibliography}
\bibliographystyle{plainnat}

\clearpage

\begin{appendices}

\crefalias{section}{appsec}
\crefalias{subsection}{appsec}
\crefalias{subsubsection}{appsec}

\setcounter{equation}{0}
\renewcommand{\theequation}{\thesection.\arabic{equation}}

\onecolumn

\section*{\LARGE Supplementary Material}
\label{sec:appendix}

\section{Experimental Details}
\label{appsec:experiment}

\paragraph{Neural Network Architecture and Optimization.}
We follow the precedents set by \citet{Sagawa2020Distributionally}, \citet{yang2023change}, and others. That is, all image datasets use a pre-trained ResNet-50 \citep{he2016resnet}, and language datasets use a pre-trained BERT \citep{devlin-etal-2019-bert}. Images are resized and cropped in the center to 224x224 pixels. For image datasets, we use SGD with momentum $0.9$. For language datasets, we use AdamW with default parameters \citep{loshchilov2017decoupled}. \Cref{tab:hyperparams-erm} shows the hyperparamenters we use for the ERM-training step and \Cref{tab:hyperparams} (last-layer) resp.~ \Cref{tab:hyperparams-all-layer} (all-layer) show the hyperparameters we use for the fine-tuning stage. We note that in the fine-tuning step with group-aware prior, we use a portion of the validation set for both terms of the loss in \Cref{eq:e-map-objective-mc}, i.e. both terms are evaluated on points from the context set, which is chosen to be the validation dataset in our case. In principle, we could use the training set for the first term in \Cref{eq:e-map-objective-mc}, as it does not require group labels, but we have chosen to evaluate both terms on the context distribution for simplicity.

\paragraph{Parameters for Ablation Studies.}
For \Cref{tab:ablation}, we use the hyperparameters detailed in \Cref{tab:hyperparams} and set the relevant parameter to its `trivial' realization. That is, we set $\rho=0$.
In this ablation study, we average our results over ten trials.
For \Cref{tab:ablation2}, we also use a `trivial' parameter choice $\gamma=1$, but also investigate results under different strengths of upweighting.
In this study, we estimate our mean and standard errors from three trials, except for when the parameter choice coincides with our parameter choice from \Cref{tab:hyperparams} (e.g. $\gamma=4$), in which case we use ten.

\begin{table}[h!]
\centering
\caption{
{\bf Table of Hyperparameter Choice for Initial ERM Finetuning.}
For all of our experiments that use a pretrained network fine-tuned on the target task with ERM, we detail the hyperparameters used. Initial LR means the inital learning rate input into the learning rate scheduler. Note that $\alpha$ is the minimum multiplier value for adjusting the learning rate for the cosine decay scheduler.
}
\label{tab:hyperparams-erm}
\begin{tblr}{
width=0.4\columnwidth,
colspec={c|ccc},
colsep=28pt,
}
\toprule
\textbf{Hyperparameter}    & \textbf{Waterbirds} & \textbf{CelebA} & \textbf{MultiNLI} \\ \hline

Epochs                     & 30                  & 5               & 3                 \\
Initial LR                 & 0.005               & 0.005           & 0.00002          \\
LR Scheduler               & Cosine Decay        & Cosine Decay    & Linear Decay      \\
$\alpha$                   & 0.01                & 0.001           & ---               \\
Batch Size                 & 128                 & 128             & 32                \\
Weight Decay               & 0                   & 0               & 0.01                 \\
\bottomrule
\end{tblr}
\end{table}

\begin{table}[h!]
\centering
\caption{
{\bf Table of Hyperparameter Choice for Last-Layer Retraining.}
For our experiments involving last-layer retraining on our group-aware prior, we detail the hyperparameters used. Epochs specifically indicate the number of epochs we re-train our last layer on, and does not include the previous ERM fine-tuning steps. Initial LR means the inital learning rate input into the learning rate scheduler. Note that $\alpha$ is the minimum multiplier value for adjusting the learning rate for the cosine decay scheduler. The parameters $\lambda$, $\gamma$, and $\rho$ are hyperparameters specific to our method.
}
\label{tab:hyperparams}
\begin{tblr}{
width=0.4\columnwidth,
colspec={c|ccc},
colsep=28pt,
}
\toprule
\textbf{Hyperparameter}    & \textbf{Waterbirds} & \textbf{CelebA} & \textbf{MultiNLI} \\ \hline

Epochs                     & 40                  & 20              & 5                 \\
Initial LR                 & 0.001               & 0.0001          & 0.00004           \\
LR Scheduler               & Cosine Decay        & Cosine Decay    & Linear Decay      \\
$\alpha$                   & 1                   & 1               & ---               \\
Batch Size                 & 128                 & 128             & 32                \\
Weight Decay               & 0                   & 0               & 0                 \\
$\lambda$                  & 1                   & 30              & 10                \\
$\gamma$                   & 4                   & 1.5             & 2                 \\
$\rho$                     & 0.15                & 0.15            & 0.1               \\
\bottomrule
\end{tblr}
\end{table}

\begin{table}[h!]
\centering
\caption{
{\bf Table of Hyperparameter Choice for All-Layer Finetuning.}
For our experiments involving all-layer finetuning on our group-aware prior, we detail the hyperparameters used. Epochs specifically indicate the number of epochs we finetune the entire network on, and does not include the previous ERM fine-tuning steps. Initial LR means the inital learning rate input into the learning rate scheduler. The parameters $\lambda$, $\gamma$, and $\rho$ are hyperparameters specific to our method.
}
\label{tab:hyperparams-all-layer}
\begin{tblr}{
width=0.4\columnwidth,
colspec={c|ccc},
colsep=28pt,
}
\toprule
\textbf{Hyperparameter}    & \textbf{Waterbirds} & \textbf{CelebA} & \textbf{MultiNLI} \\ \hline

Epochs                     & 40                  & 10              & 1                 \\
Initial LR                 & 0.001               & 0.0001          & 0.000005          \\
LR Scheduler               & Linear Decay        & Linear Decay    & Linear Decay      \\
Batch Size                 & 128                 & 128             & 32                \\
Weight Decay               & 0                   & 0               & 0                 \\
$\lambda$                  & 15                  & 200             & 10                \\
$\gamma$                   & 4                   & 4               & 4                 \\
$\rho$                     & 0.15                & 0.1             & 0.1               \\
\bottomrule
\end{tblr}
\end{table}

\clearpage

\section{Method Details}

We define a general family of group-aware priors by specifying a Bernoulli observation model
\begin{align}
\begin{split}
    \hat{p}(\hat{z} = 1 \vbar \theta ; f, p_{\smash{\hat{X}, \hat{Y}}})
    &
    =
    \exp(- \lambda \mathbb{E}_{p_{\smash{\hat{X}, \hat{Y}}}}[c(\hat{X}, \hat{Y}, \theta) ] )
    \\
    \hat{p}(\hat{z} = 0 \vbar \theta ; f, p_{\smash{\hat{X}, \hat{Y}}})
    &
    =
    1 - \hat{p}(\hat{z} = 1 \vbar \theta ; f, p_{\smash{\hat{X}, \hat{Y}}}) ,
\end{split}
\end{align}
where $c : \mathcal{X} \times \mathcal{Y} \times \mathbb{R}^{P} \rightarrow \mathbb{R}_{\geq}$ is a `cost' function and $\lambda > 0$ is a scaling parameter.

To define a specific member of this family of priors, we specify a cost function
\begin{align}
    c(\hat{x}, \hat{y}, \theta)
    \defines
    \ell(\hat{y}, f(\hat{x} ; \theta + \rho \epsilon(\theta, \hat{x}, \hat{y}))) 
\end{align}
where $\rho > 0$ is a scaling parameter, $\ell$ is a cross-entropy loss function for classification tasks and a mean-squared error loss function for regression tasks, and $\epsilon(\theta, \hat{x}, \hat{y})$ is a worst-case perturbation to the model parameters proposed in \citet{foret2021sharpnessaware} and given by
\begin{align}
    \epsilon(\theta, \hat{x}, \hat{y})
    \defines
    \perp \frac{\nabla_{\theta} \ell(\hat{y}, f(\hat{x} ; \theta)) }{ \| \nabla_{\theta} \ell(\hat{y}, f(\hat{x} ; \theta)) \|_{2}} ,
\end{align}
where $\perp$ is the \texttt{stop\_gradient} operator.

It is worth briefly noting the implications of using the \texttt{stop\_gradient} operator in a prior probability density function.
Since the perturbation $\epsilon(\theta, \hat{x}, \hat{y})$ is a function of $\theta$, it would normally be differentiable with respect to $\theta$, which would affect the gradient of the robustness regularization term in the final objective given in \Cref{eq:e-map-objective-mc}.
Since we apply the \texttt{stop\_gradient} operator, $\epsilon(\theta, \hat{x}, \hat{y})$ is treated as a constant for the purposes of gradient computation.
Since the gradients of the robustness regularization term with respect to $\theta$ are therefore different when the \texttt{stop\_gradient} operator is applied, this implies that the application of the \texttt{stop\_gradient} operator implicitly changes the prior probability density function.
To see how the application of the \texttt{stop\_gradient} operator changes the prior probability density function, we first note that the gradient of the cost function {\em without} the \texttt{stop\_gradient} operator in $\epsilon(\theta, \hat{x}, \hat{y})$, evaluated at parameters $\theta_{t}$, is given by
\begin{align}
    \nabla_{\theta} \ell(\hat{y}, f(\hat{x} ; \theta + \rho \tilde{\epsilon}(\theta)))|_{\theta=\theta_{t}}
    &
    =
    \frac{\dee (\theta + \tilde{\epsilon}(\theta))}{\dee \theta}\bigg|_{\theta = \theta_{t}} \nabla_{\theta} \ell(\hat{y}, f(\hat{x} ; \theta))|_{\theta = \theta_{t} + \tilde{\epsilon}(\theta_{t})}
    \\
    &
    =
    \left.\nabla_{\theta} \ell(\hat{y}, f(\hat{x} ; \theta))\right|_{\theta = \theta_{t} + \tilde{\epsilon}(\theta_{t})}
    +
    \frac{\dee  \tilde{\epsilon}(\theta)}{\dee \theta}\bigg|_{\theta = \theta_{t}} \nabla_{\theta} \ell(\hat{y}, f(\hat{x} ; \theta))|_{\theta = \theta_{t} + \tilde{\epsilon}(\theta_{t})} ,
\end{align}
where
\begin{align}
\SwapAboveDisplaySkip
    \tilde{\epsilon}(\theta)
    \defines
    \frac{\nabla_{\theta} \ell(\hat{y}, f(\hat{x} ; \theta)) }{ \| \nabla_{\theta} \ell(\hat{y}, f(\hat{x} ; \theta)) \|_{2}} .
\end{align}
In contrast, {\em with} the \texttt{stop\_gradient} operator in $\epsilon(\cdot)$, the gradient of the cost function, evaluated at $\theta_{t}$ is given by
\begin{align}
    \nabla_{\theta} \ell(\hat{y}, f(\hat{x} ; \theta + \rho \epsilon(\theta, \hat{x}, \hat{y})))|_{\theta = \theta_{t}}
    =
    \left.\nabla_{\theta} \ell(\hat{y}, f(\hat{x} ; \theta))\right|_{\theta = \theta_{t} + \epsilon(\theta_{t})} ,
\end{align}
that is, it does not contain the term $\smash{\frac{\dee \tilde{\epsilon}(\theta)}{\dee \theta}|_{\theta = \theta_{t}} \nabla_{\theta} \ell(\hat{y}, f(\hat{x} ; \theta))|_{\theta = \theta_{t} + \tilde{\epsilon}(\theta_{t})}}$.

In order to obtain this gradient when using the perturbation function $\tilde{\epsilon}(\theta)$ (which does not use the \texttt{stop\_gradient} operator), we need to include an additive term in the cost function whose gradient is equal to $\smash{\frac{\dee  \tilde{\epsilon}(\theta)}{\dee \theta} \nabla_{\theta} \ell(\hat{y}, f(\hat{x} ; \theta))}$.
To obtain this term, all we need to do is to take the integral of $\smash{\frac{\dee  \tilde{\epsilon}(\theta)}{\dee \theta} \nabla_{\theta} \ell(\hat{y}, f(\hat{x} ; \theta))}$ with respect to $\theta$:
\begin{align}
    A(\theta)
    \defines
    - \int \frac{\dee  \tilde{\epsilon}(\theta)}{\dee \theta} \nabla_{\theta} \ell(\hat{y}, f(\hat{x} ; \theta)) \dee \theta .
\end{align}
This integral exists for all evaluation points for which $\tilde{\epsilon}(\cdot)$ and $ \ell(\hat{y}, f(\hat{x} ; \cdot))$ are differentiable with respect to $\theta$.
While this integral may be analytically intractable, we do not need to compute it unless we want to compute the value of the prior probability density for a given parameter configuration, and since we only need the prior density for optimization, we can simply use the \texttt{stop\_gradient} operator to compute the desired gradients.

\clearpage

\section{Dataset Details}

Below, we provide information about each of the datasets used in our empirical evaluation.
All images used in this manuscript are illustrative only (i.e., they differ from the actual images in the datasets used in this paper) and are taken from \url{https://unsplash.com} (for birds) and \url{https://www.istockphoto.com} (for people) under unlimited, perpetual, nonexclusive, worldwide license.

\subsection{Vision Data}

\setlength{\tabcolsep}{4.5pt}
\begin{table}[h!]
\centering
\caption{
    {\bf Summary of Vision Dataset Properties.}
}
\label{tab:datasets_vision}
\begin{tabular}{lcccc}
\toprule
\multicolumn{5}{c}{\textbf{Waterbirds}}                                   \\
\hline
\textbf{}               & $g_1$  & $g_2$   & $g_3$   & $g_4$    \\
  \begin{tabular}{l}
    \hspace*{-4pt}\textbf{Example Image}\\[0.06\textheight]
  \end{tabular}  & \includegraphics[height=0.07\textheight]{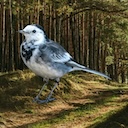}   & \includegraphics[height=0.07\textheight]{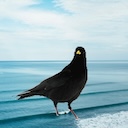} & \includegraphics[height=0.07\textheight]{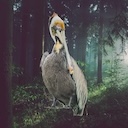} & \includegraphics[height=0.07\textheight]{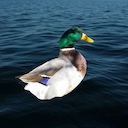} \\[-0.03\textheight]
\textbf{Group Description}    & Landbird on Land & Landbird on Water & Waterbird on Land & Waterbird on Water \\
\textbf{Class Label}    & 0                & 0                 & 1                 & 1                  \\
\textbf{Attribute Label}    & 0                & 1                 & 0                 & 1                  \\
\textbf{Group Label}    & 0                & 1                 & 2                 & 3                  \\
\textbf{\# Training Data}  & 3,498             & 184               & 56                & 1,057               \\
\textbf{\# Validation Data}   & 467              & 466               & 133               & 133                \\
\textbf{\# Test Data}   & 2,255              & 2,255               & 642               & 642                \\
\toprule
\multicolumn{5}{c}{\textbf{CelebA}}                                  \\
\hline 
\textbf{}               & $g_1$  & $g_2$   & $g_3$   & $g_4$    \\
  \begin{tabular}{l}
    \hspace*{-4pt}\textbf{Example Image}\\[0.13\textheight]
  \end{tabular}  & \includegraphics[width=0.15\textwidth]{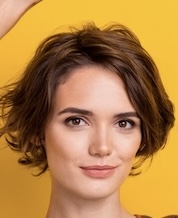} & \includegraphics[width=0.15\textwidth]{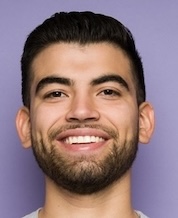} &\includegraphics[width=0.15\textwidth]    {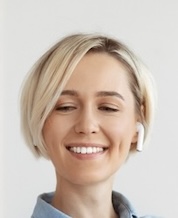}&\includegraphics[width=0.15\textwidth]{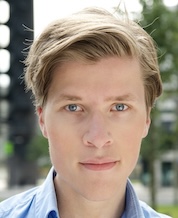}     \\[-0.065\textheight]
\textbf{Group Description}    & Non-blonde Woman & Non-blonde Man    & Blonde Woman      & Blonde Man         \\
\textbf{Class Label}    & 0                & 0                 & 1                 & 1                  \\
\textbf{Attribute Label}    & 0                & 1                 & 0                 & 1                  \\
\textbf{Group Label}    & 0                & 1                 & 2                 & 3                  \\
\textbf{\# Training Data}  & 71,629            & 66,874             & 22,880             & 1,387               \\
\textbf{\# Validation Data}   & 8,535             & 8,276              & 2,874              & 182                \\
\textbf{\# Test Data}   & 9,767              & 7,535               & 2,480               & 180                \\
\bottomrule
\end{tabular}
\end{table}

\clearpage

\subsection{Language Data}

\begin{table}[h!]
\centering
\caption{
    {\bf Summary of Language Dataset Properties.}
}
\label{tab:datasets_language}
\begin{tblr}{width=\textwidth,
colspec={X[l]X[c]X[c]X[c]},
rowspec={Q[m]Q[m]Q[m]Q[m]Q[m]Q[m]Q[m]Q[m]Q[m]Q[m]Q[m]Q[m]Q[m]Q[m]Q[m]Q[m]Q[m]Q[m]Q[m]},
cell{1}{1} = {c=4}{c},
cell{3}{1} = {r=2}{},
cell{12}{1} = {r=2}{},
}
\toprule
\textbf{Multi-Genre Natural Language Inference (MultiNLI) corpus} & & & \\
\hline
\textbf{}               & $g_1$  & $g_2$   & $g_3$ \\
\textbf{Example Text}
& (P): \textit{if residents are unhappy, they can put wheels on their homes and go someplace else, she said.} 
& (P): \textit{within this conflict of values is a clash about art.}
& (P): \textit{there was something like amusement in the old man's voice.}
\\
& (H): \textit{residents are stuck here but they can't go anywhere else.} 
& (H): \textit{there is no clash about art.} 
& (H): \textit{the old man showed amusement.} 
\\
\textbf{Group Description}    & Contradiction without Negations & Contradiction with Negations & Entailment without Negations  \\
\textbf{Class Label}    & 0                & 0                 & 1                            \\
\textbf{Attribute Label}& 0                & 1                 & 0                      \\
\textbf{Group Label}    & 0                & 1                 & 2                  \\
\textbf{\# Training Data}  & 57,498 & 11,158 & 67,376            \\
\textbf{\# Validation Data}   & 22,814 & 4,634 & 26,949          \\
\textbf{\# Test Data}   & 34,597 & 6,655 & 40,496          \\
\hline\\[-20pt]
               & $g_4$ & $g_5$ & $g_6$
\\
\vspace*{-20pt}\textbf{Example Text} 
& (P): \textit{in 1988, the total cost for the postal service was about \$36.} 
& (P): \textit{yeah but even even cooking over an open fire is a little more fun isn't it.} 
& (P): \textit{that’s not too bad.} 
 \\
& (H): \textit{the postal service cost us citizens almost nothing in the late 80's.} 
& (H): \textit{i like the flavour of the food.} 
& (H): \textit{it’s better than nothing.} 
 \\
\textbf{Group Description}    & Entailment with Negations & Neutral without Negations & Neutral with \hspace{50pt}Negations \\
\textbf{Class Label}    & 1  & 2 & 2                \\
\textbf{Attribute Label}             & 1  & 0 & 1                \\
\textbf{Group Label}               & 3  & 4 & 5               \\
\textbf{\# Training Data}   & 1,521 & 66,630 & 1,992               \\
\textbf{\# Validation Data}   & 613 & 26,655 & 797          \\
\textbf{\# Test Data}   & 886 & 39,930 & 1,148          \\
\bottomrule
\end{tblr}
\end{table}

\end{appendices}

\end{document}